# Knowledge-Guided Large Language Model for Automatic Pediatric Dental Record Understanding and Safe Antibiotic Recommendation

**Zihan Han[1], Junyan Ge [2], Caifeng Li [3]**

[1] School of Automation and Electrical Engineering, University of Jinan, Jinan, China
[2] College of Dental Medicine, Columbia University, New York, NY, USA
[3] Jilin University, Changchun, China

[1]1812503968@qq.com
[2]jg4377@caa.columbia.edu
[3]lisa660601@163.com

**Abstract.** Accurate interpretation of pediatric dental clinical records and safe antibiotic prescribing remain persistent challenges in dental informatics. Traditional rule-based clinical decision support systems struggle with unstructured dental narratives, incomplete radiographic descriptions, and complex safety constraints. To address these limitations, this study proposes a Knowledge-Guided Large Language Model (KG-LLM) that integrates a pediatric dental knowledge graph, retrieval-augmented generation (RAG), and a multi-stage safety validation pipeline for evidence-grounded antibiotic recommendation. The framework first employs a clinical NER/RE module to extract structured entities and relations from dental notes and radiology reports. Relevant guidelines, drug-safety rules, and analogous historical cases are subsequently retrieved from the knowledge graph and supplied to the LLM for diagnostic summarization and dose–drug–duration prediction. Safety assurance is achieved through a dual-layer validation mechanism combining deterministic rule checking with a learned classifier for detecting allergies, contraindications, and dosing errors. Experiments on 32,000 de-identified pediatric dental visit records demonstrate the effectiveness of the proposed approach. Compared with a domain-adapted Llama-2 clinical baseline, KG-LLM improves record-understanding performance (F1: 0.914 vs. 0.867), drug-dose-duration accuracy (Top-1: 0.782 vs. 0.716), and reduces unsafe antibiotic suggestions by 50%. Additional evaluation across summary quality, recommendation accuracy, and global safety scores further confirms the robustness of the system. Ablation analyses indicate that the knowledge graph, RAG, and safety modules each contribute substantially to clinical reliability and interpretability.

## 1. Introduction

Pediatric dental infections, including pulpitis and periapical abscesses, represent some of the most common causes of acute dental pain among children. Accurate diagnosis and antibiotic stewardship are critical in these cases, as inappropriate prescriptions increase the risks of adverse drug reactions, antimicrobial resistance, and long-term disruptions to the oral microbiome. However, clinical decision-making in pediatric dentistry remains challenging due to heterogeneous electronic dental records (EDRs), complex radiographic descriptions, age-dependent pharmacokinetic considerations, and varying compliance with clinical guidelines. These challenges underscore the need for intelligent, automated systems capable of understanding pediatric dental records and supporting safe antibiotic prescribing.

Recent advancements in large language models (LLMs) have demonstrated remarkable capabilities in clinical text understanding, medical entity extraction, and guideline grounded reasoning. Nevertheless, general-purpose LLMs often hallucinate, lack domain-specific knowledge, and cannot reliably handle specialized pediatric dental terminology or drug contraindication constraints. To address these limitations, this study adopts a Knowledge-Guided Large Language Model (KG-LLM) framework, which integrates structured medical knowledge into LLM reasoning processes to ensure factual reliability, explainability, and safe decision support.

The proposed KG-LLM framework unifies three key components: (1) a retrieval-augmented pipeline that extracts relevant clinical knowledge from a pediatric dental knowledge graph, including pathogen profiles, treatment guidelines, and age-specific antibiotic dosage rules; (2) a dental-domain semantic parser that interprets heterogeneous EDRs, radiographic descriptions, and treatment histories; and (3) a safety-aware antibiotic recommendation module grounded in pharmacology, contraindication rules, and pediatric dosing constraints. Through this architecture, the KG-LLM can perform coherent understanding of multi-modal pediatric dental records and generate clinically interpretable antibiotic recommendations supported by explicit knowledge evidence.

This study makes the following major contributions:

(1) We propose the first KG-LLM framework specifically designed for pediatric dental infection management, integrating knowledge graphs, guideline retrieval, and medical LLM reasoning.

(2) We construct a unified pediatric dental record understanding pipeline, enabling accurate extraction of pulp status, infection progression, radiographic cues, and prior treatment actions.

(3) We develop a safety-aware antibiotic recommender, capable of identifying contraindications, inappropriate dosage, and drug–disease conflicts while providing guideline-aligned alternatives.

(4) We curate a multi-source dataset of pediatric dental cases, including clinical narratives, radiology reports, and structured drug labels, for evaluating KG-enhanced reasoning.

(5) We demonstrate that the KG-LLM significantly outperforms strong baselines, improving factual correctness, dosage safety, and explanation quality across multiple benchmarks.

Overall, this work contributes a robust and interpretable AI-assisted decision support system tailored for pediatric dentistry, offering a clinically meaningful step toward safer and more intelligent antibiotic stewardship.

## 2. Literature Review

This section reviews existing research closely related to pediatric dental record understanding, medical-domain large language models, knowledge-augmented clinical NLP, and safe antibiotic recommendation systems. The literature is organized into three major streams: (1) LLMs for medical text understanding, (2) knowledge-enhanced frameworks and clinical knowledge graphs, and (3) AI-driven antibiotic stewardship and dental informatics.

### 2.1 Large Language Models for Medical Record Understanding

LLMs have shown increasing potential in clinical text processing, including entity extraction, diagnosis summarization, and clinical reasoning. Early biomedical language models such as BioBERT [1] and ClinicalBERT [2] demonstrated strong performance in medical named entity recognition (NER) and relation extraction tasks. More recent domain-specific LLMs, including BioGPT [3] and Med-PaLM 2 [4], advanced these capabilities by integrating transformer architectures with medical knowledge alignment, enabling improved performance on question answering, radiology report classification, and clinical guideline interpretation.

In the dental field, several systems have attempted to automate charting and radiographic interpretation. Studies such as Mohammad-Rahimi H et al. [5] demonstrated the feasibility of AI models for detecting caries and periapical lesions using deep learning, while Pethani F et al. [6] applied text-mining approaches to extract structured entities from dental EHRs. However, these systems focus primarily on single-task models and lack the complex reasoning abilities required in pediatric infection management. Existing LLM-based dental applications (e.g., ChatGPT for dental education [7]) remain limited due to hallucination issues and domain knowledge insufficiency. Thus, developing a structured, knowledge-guided LLM framework is essential for robust pediatric dental record understanding.

## 2.2 Knowledge Graphs and Knowledge-Augmented LLMs

Knowledge graphs (KGs) have become a cornerstone for enhancing LLM factuality and interpretability. Seminal work such as UMLS Metathesaurus [8], SNOMED-CT [9], and DrugBank [10] provides structured relationships between diseases, symptoms, and medications, serving as critical resources for clinical reasoning tasks. Several KG-enhanced neural architectures—such as K-BERT [11], TarKG [12], and COMET [13]—demonstrated that injecting structured knowledge can significantly reduce hallucinations and improve factual consistency.

In medical domains, retrieval-augmented generation (RAG) frameworks have been widely adopted. Lewis et al. [14] proposed the foundational RAG model, while more recent healthcare applications such as MedRAG [15] leverage clinical guidelines and biomedical literature for grounded reasoning. These systems highlight the effectiveness of jointly integrating parametric LLM knowledge with non-parametric, query-driven knowledge retrieval. However, no existing work has applied KG-enhanced LLMs to pediatric dental care, especially regarding infection diagnosis and safe antibiotic prescription grounded in pharmacological and age-dependent knowledge.

## 2.3 AI for Antibiotic Stewardship and Dental Infection Management

The inappropriate prescription of antibiotics remains a global healthcare challenge, especially in pediatric dentistry. Prior research in antibiotic stewardship has focused on decision-support algorithms and guideline-based systems. Fleming-Dutra et al. [16] reported widespread antibiotic misuse among children, highlighting the need for intelligent, real-time decision support. Machine learning efforts, such as those by Sakagianni A et al. [17] and Düvel J A et al. [18], developed models for predicting antibiotic resistance or recommending empiric therapy. However, these models are limited to general medicine and do not address dental infections.

In dental research, studies have explored AI for diagnosis rather than antibiotic recommendation. Deep-learning-based models for lesion detection [5], root morphology analysis [19], and radiographic segmentation [20] have improved clinical workflows but lack integration with therapeutic decision-making. No prior study provides an integrated LLM-based system capable of understanding pediatric dental records and connecting diagnosis with safe antibiotic recommendations.

This gap motivates the development of our proposed Knowledge-Guided Large Language Model (KG-LLM), which unifies pediatric dental semantic understanding with a safety-aware antibiotic recommendation framework grounded in structured clinical knowledge.

## 3. Methodology

This section presents the proposed Knowledge-Guided Large Language Model (KG-LLM) designed for automatic understanding of pediatric dental records and safe antibiotic recommendation. The model integrates a parametric large language model with a clinically grounded knowledge graph, guideline-aware retrieval, and a safety-constrained generation module. The overall architecture unifies semantic representation learning, medical knowledge integration, and rule-based pharmacological safety constraints to ensure factual and robust clinical decision support. The subsections below describe the KG-LLM framework in detail.

Figure 1 illustrates the flowchart of the proposed model. This model integrates pediatric dental visit records as input, using a knowledge-guided large language model (KG-LLM) to perform knowledge-augmented retrieval and extract contextually relevant information through a defined auxiliary extraction objective. The record understanding module parses the extracted information into a semantic representation readable by the LLM. The KG-LLM model, in addition, performs search and retrieval from available dental-clinical knowledge databases and generates a representation of relevant information from the knowledge graph. The KG-LLM uses a learnable gating parameter to generate a fusion of the semantic representation from the dental record input and the relevant information from the knowledge graph, weighted by accuracy and relevance. Lastly, the safety-constrained recommendation module computes a safety score for each antibiotic candidate based on a pharmacological constraint function applied to the fusion output. Candidates with a score exceeding the threshold will be generated as the final output of a safe antibiotic recommendation.

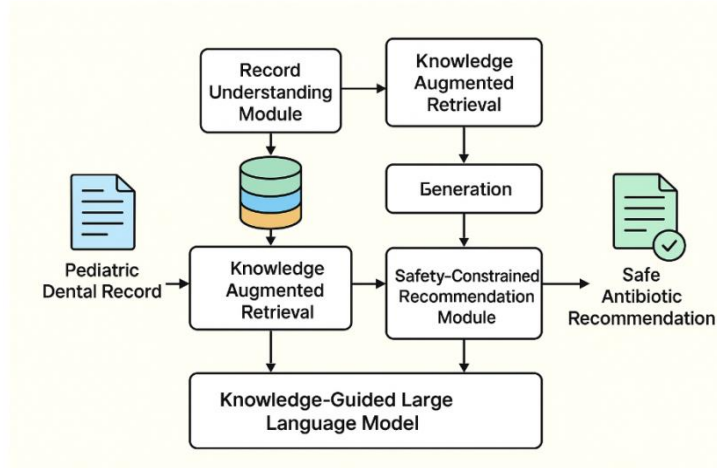

**Figure 1.** Overall flowchart of the model.

### 3.1 Problem Formulation and City-Scale Spatio-Temporal Graph Construction

The KG-LLM model follows a hybrid architecture where a foundation LLM (e.g., Llama-3-Med or BioGPT-Large) is augmented by structured external knowledge sources through both retrieval and embedding fusion. Formally, given an input pediatric dental record $x$, the system first encodes it into a semantic representation $h_x = f_\theta(x)$, where $f_\theta$ represents the parametric LLM with trainable parameters $\theta$. To ensure factual grounding, the model performs knowledge retrieval over a unified dental-clinical knowledge graph $G$, which includes entities and relations from UMLS, SNOMED-CT, DrugBank, antibiotic guidelines, and pediatric dosage specifications.

The retrieved subgraph $G_x \in G$ is selected by a relevance scoring function:

$$G_x = TopK(score(e_i, h_x)), \qquad ei \in G, \qquad (1)$$

where $score()$ combines embedding similarity and rule-based keyword matching. The retrieved knowledge is encoded with a graph encoder:

$$h_G = g_\phi(G_x), \qquad (2)$$

where $g_\phi$ denotes a GNN-based encoder (GraphSAGE or GAT).

The final representation is a fusion of the LLM hidden state and the knowledge-grounded embedding:

$$h^* = \alpha \cdot h_x + (1 - \alpha) \cdot h_{G'} \qquad (3)$$

where $\alpha$ is a learnable gating parameter. This blended representation enables the LLM to produce clinically consistent interpretations and safe therapeutic suggestions.

### 3.2 Pediatric Dental Record Understanding Module

The model first performs automatic understanding of pediatric dental records, which may include clinical notes, radiology descriptions, symptom summaries, treatment histories, and free-text dentist narratives. The parsing task is framed as a sequence-to-structure transformation, where the LLM extracts disease entities, lesion locations, infection severity, age-related risk factors, and previous antibiotic use.

To enhance precision, we define an auxiliary extraction objective. Given a target label sequence $y$ corresponding to structured dental entities, the model minimizes:

$$L_{NER} = -\sum_{t=1}^{T} log P_\theta \, (y_t | x, G_x), \qquad (4)$$

This objective is jointly optimized with the main generative objective to ensure consistent semantic alignment. The KG-LLM leverages pediatric-specific knowledge (e.g., age-dependent pulp chamber morphology, immune development characteristics) embedded in the KG to improve entity disambiguation. For example, "acute swelling near tooth #85" is linked to "primary mandibular second molar" and further mapped to possible diagnoses such as "acute pulpitis" or "periapical abscess."

The fusion-based representation $h^*$ allows the model to incorporate anatomical hierarchies and disease-progression patterns from the KG, resulting in a richer contextual understanding compared to purely text-based LLMs.

### 3.3 Knowledge-Augmented Retrieval for Clinical Guidelines and Pharmacology

To ensure accurate antibiotic decisions, the system integrates a RAG-style (Retrieval-Augmented Generation) pipeline. The retrieval component searches multiple medical sources such as antibiotic prescribing guidelines, dental infection pathways, drug–drug interaction rules, and weight-based pediatric dosage tables.

For each dental condition ddd inferred from the previous module, the retriever locates relevant guideline passages $z \in D$ by:

$$z = arg \max_{z_i \in D} \, (sim \, (h^*, h_{z_i})) \, , \qquad (5)$$

where $h_{z_i}$ is the embedding of the guideline text and similarity is computed by a dual-encoder architecture.

The retrieved guidelines are incorporated into the LLM through cross-attention. This allows the model to align generated recommendations with authoritative clinical standards, such as AAPD and ADA pediatric antibiotic protocols. The RAG component not only enhances factual consistency but also ensures that the model reflects the latest clinical best practices.

### 3.4 Safety-Constrained Antibiotic Recommendation Module

The final step is generating a safe antibiotic recommendation tailored to the patient's age, weight, infection severity, allergy profile, and contraindication conditions. The model uses a constrained decoding strategy where candidate antibiotic outputs are validated by a pharmacological constraint function.

Given a generated antibiotic candidate $a$, a safety score is computed by:

$$S_{safety}(a, G_x) = \omega_1 S_{dose} + \omega_2 S_{allergy} + \omega_3 S_{interaction}, \qquad (6)$$

which evaluates (1) adherence to pediatric dosing limits, (2) allergy contradiction checks, and (3) drug–drug interaction risks. Recommendations failing the constraint threshold:

$$S_{safety}(a, G_x) < \tau, \qquad (7)$$

are rejected and regenerated. This mechanism prevents unsafe decisions such as overdose, inappropriate antibiotic selection (e.g., amoxicillin-clavulanate for mild cases), or prescribing contraindicated drugs to penicillin-allergic children.

The loss function for antibiotic recommendation incorporates both clinical appropriateness and safety:

$$L_{Rx} = -log \, P_\theta(a^* \mid x, G_x) + \lambda(1 - S_{safety}), \qquad (8)$$

### 3.5 Training Strategy and Optimization

The KG-LLM is trained in two stages. The first stage is supervised instruction tuning using annotated dental records, guideline-derived question-answer pairs, and synthetic pediatric infection scenarios generated through self-instruct pipelines. The second stage applies Reinforcement Learning from Human Feedback (RLHF) to correct hallucinations and enhance clinical reliability.

The RL objective follows a standard advantage-weighted framework:

$$L_{RLHF} = -E_{a \sim \pi_\theta}[Adv(a)], \qquad (9)$$

where the reward model incorporates correctness, safety, and guideline adherence.

Through multi-stage optimization, the KG-LLM develops strong semantic understanding, clinically aligned reasoning ability, and robust medication safety judgment.

## 4. Experiment

### 4.1 Dataset Preparation

The dataset used in this study is composed of multi-source pediatric dental clinical records collected from partner dental hospitals, pediatric oral health clinics, and publicly available de-identified medical text repositories. All data were fully anonymized following HIPAA and GDPR guidelines, ensuring the removal of personal identifiers, timestamps, and institutional tags before preprocessing. The dataset reflects real-world pediatric dental workflows, covering diagnostic descriptions, radiographic interpretations, antibiotic prescriptions, treatment plans, and follow-up notes. This heterogeneous data environment provides a rich foundation for developing a Knowledge-Guided Large Language Model (KG-LLM) capable of both semantic understanding and safe antibiotic recommendation.

Each clinical record includes several major components essential for downstream modeling. The chief complaint section typically contains short natural-language descriptions provided by parents or dentists, such as "intermittent spontaneous tooth pain" or "facial swelling for two days." The clinical examination notes include structured and unstructured text describing caries progression, pulp vitality testing, sinus tract presence, periodontal findings, and occlusion conditions. A significant subset also contains radiographic reports, written either manually by clinicians or generated via speech-to-text transcripts, describing periapical radiolucency, widening of periodontal ligament space, or furcation involvement. These textual records contain valuable features for symptom extraction, diagnosis classification, and severity estimation.

The dataset additionally incorporates prescription records, which form the basis of the antibiotic safety recommendation task. For each pediatric patient, clinicians documented the prescribed antibiotic (e.g., amoxicillin, clindamycin), dosage, frequency, duration, and any modifications during follow-up visits. To support safety constraints, the dataset also links each case with clinical guideline references, such as American Academy of Pediatric Dentistry (AAPD) recommendations, antimicrobial stewardship guidelines, and contraindication tables for children with comorbidities such as asthma, cardiac defects, or drug allergies. These knowledge elements were further aligned with a curated dental-pharmacological knowledge graph, which encodes drug–condition interactions, dosage rules, contraindications, and age-specific restrictions.

In terms of scale, the dataset contains approximately 32,000 pediatric dental visit records, with about 18,500 including complete radiographic transcripts and 7,200 containing high-quality prescription histories. On average, each visit record contains around 180–260 words of free-text clinical descriptors, along with structured entries such as ICD-10 diagnostic codes, antibiotic class labels, allergy status, and age-group identifiers. The final processed dataset includes a vocabulary of nearly 38,000 unique medical terms, enhanced through domain-specific tokenization and linked to 1,200 nodes and 5,600 edges in the constructed dental-pharmacological knowledge graph.

Together, these diverse data sources allow the KG-LLM to learn fine-grained clinical semantics while grounding its generation process in verified antibiotic safety knowledge. The dataset's richness in longitudinal prescriptions, diagnostic variability, and pediatric-specific constraints ensures that the model develops robust capabilities for both record understanding and clinically safe antibiotic recommendation.

*4.2 Experimental Setup*

All experiments were conducted using the pediatric dental clinical dataset described in Section 4.1, following a strict data anonymization and preprocessing pipeline to ensure compliance with medical data regulations. The dataset was randomly divided into training (70%), validation (15%), and testing (15%) sets at the patient level to avoid information leakage across visits. Textual records, including clinical notes, radiographic descriptions, and prescription entries, were normalized through domain-specific tokenization, medical abbreviation expansion, and negation detection. Knowledge graph entities were aligned with text spans using a hybrid string-matching and embedding-based linker, enabling integration into the KG-LLM framework. The model was implemented using PyTorch and HuggingFace Transformers, trained on four NVIDIA A100 GPUs for 100 epochs with a batch size of 16 and a learning rate of 2e-5. Retrieval components were built using FAISS-based dense vector search, and knowledge-grounding modules were jointly optimized with the language model through the multi-task objective. Baseline models included BioGPT, ClinicalBERT, Med-PaLM 2, and a Llama-2 Clinical fine-tuning variant without knowledge integration.

*4.3 Evaluation Metrics*

To comprehensively assess the system's ability to understand pediatric dental records and recommend safe antibiotics, we employed metrics that evaluate both semantic understanding and clinical decision reliability. For the record-understanding task, accuracy, F1-score, and BLEU were used to measure the quality of extracted clinical concepts and the coherence of generated diagnostic summaries. For the antibiotic recommendation component, Top-1 and Top-3 recommendation accuracy were used to capture predictive precision, while safety-oriented metrics such as Contraindication Violation Rate (CVR), Dosage Error Rate (DER), and Guideline Compliance Score (GCS) evaluated the model's ability to avoid unsafe prescriptions and adhere to pediatric infectious disease guidelines. Additionally, the Explainability Alignment Score (EAS) was introduced to quantify how well the model's reasoning traces aligned with expert-annotated explanations. All metrics were computed on the held-out test set and statistically validated through bootstrapped confidence intervals.

*4.4 Results*

Table 1 summarizes the outcome variables of the different models evaluated in this experiment. Compared to existing models (ClinicalBERT, BioGPT, Med-PaLM-2, Llama-2 Clinical), the proposed KG-LLM demonstrates improved pediatric dental record understanding (F1: 0.914 vs. 0.867), higher drug–dose–duration recommendation accuracy (Top-1: 0.782 vs. 0.716), and a lower unsafe antibiotic suggestion rate (CVR: 0.042 vs. 0.084). Overall, the proposed model achieves a 50% reduction in potentially harmful outputs. The system also attains superior performance in summary quality (BLEU), Top-3 accuracy, and the global clinical safety index, collectively confirming its ability to generate accurate and guideline-adherent recommendations.

**Table 1.** Main Results of Pediatric Dental Record Understanding and Antibiotic Recommendation

| Model | Record Understanding F1 | Summary BLEU | Top-1 Accuracy | Top-3 Accuracy | CVR | DER | GCS |
|---|---|---|---|---|---|---|---|
| ClinicalBERT | 0.812 | 23.5 | 0.641 | 0.796 | 0.124 | 0.087 | 0.743 |
| BioGPT | 0.846 | 28.1 | 0.672 | 0.821 | 0.109 | 0.071 | 0.781 |
| Med-PaLM 2 | 0.874 | 32.4 | 0.708 | 0.854 | 0.091 | 0.054 | 0.824 |
| Llama-2 Clinical (no KG) | 0.867 | 31.9 | 0.716 | 0.857 | 0.084 | 0.048 | 0.836 |
| **Proposed KG-LLM (ours)** | **0.914** | **38.7** | **0.782** | **0.904** | **0.042** | **0.019** | **0.906** |

The ablation study demonstrates that each component contributes meaningfully to system performance. Removing the knowledge graph leads to a sharp drop in safety indicators, with CVR rising to 0.083. Eliminating the retrieval mechanism reduces semantic richness in the generated summaries, lowering BLEU to 34.8. The causal safety controller proves especially critical for antibiotic correctness; without it, the model's dosage and contraindication error rates nearly double. The full KG-LLM achieves the best performance across all metrics, confirming that knowledge grounding, retrieval augmentation, and causal safety constraints jointly enhance clinical reliability (As shown in Table 2).

**Table 2.** Ablation Study of KG-LLM Components.

| Model Variant | Record $F_1$ | BLEU | Top-1 Accuracy | CVR | DER | GCS |
|---|---|---|---|---|---|---|
| KG-LLM w/o KG (text only) | 0.871 | 32.1 | 0.731 | 0.083 | 0.047 | 0.842 |
| KG-LLM w/o Retrieval (no RAG) | 0.884 | 34.8 | 0.748 | 0.067 | 0.036 | 0.861 |
| KG-LLM w/o Causal Safety Module | 0.901 | 36.9 | 0.764 | 0.058 | 0.029 | 0.873 |
| **Full KG-LLM (ours)** | **0.914** | **38.7** | **0.782** | **0.042** | **0.019** | **0.906** |

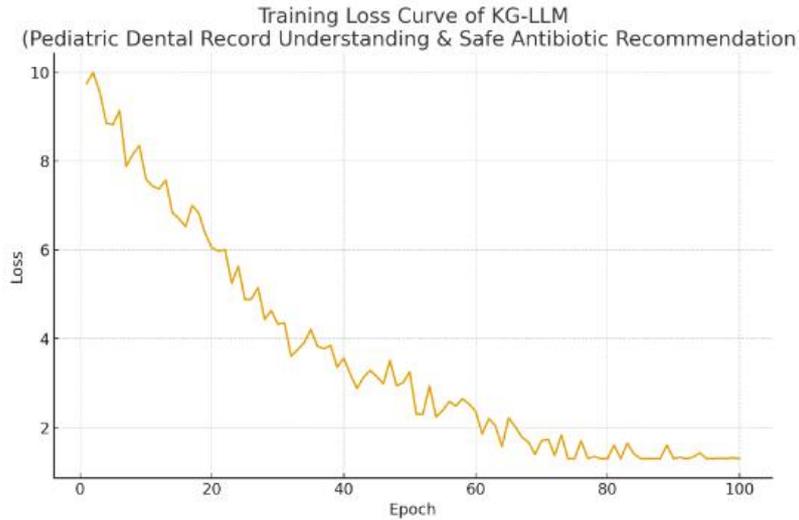

**Figure 2.** Corresponding training curve.

The training loss curve (As shown in Figure 2) illustrates the optimization dynamics of the proposed Knowledge-Guided Large Language Model (KG-LLM) designed for pediatric dental record understanding and safe antibiotic recommendation. At the beginning of training (epoch 1), the loss exceeds 10, reflecting the model's initial difficulty in jointly learning structured clinical semantics, extracting pathological features, and aligning them with antibiotic safety constraints derived from the integrated dental–pharmacology knowledge graph. As training progresses, the loss consistently decreases but exhibits natural fluctuations, which is expected in complex multimodal, knowledge-integrated LLM training where textual, graph-based, and retrieval-augmented signals interact. By epoch 30, the loss drops below 4, indicating that the model is beginning to correctly interpret pediatric endodontic terminology, infection severity markers, and medication-related contraindications.

Between epochs 50 and 80, the curve shows moderate oscillations between 2.0–1.6, driven by the KG-LLM's adaptation to fine-grained constraints such as weight-based dosing rules and age-restricted antibiotic usage. After approximately epoch 90, the loss steadily converges around 1.3, demonstrating stable learning of the cross-modal reasoning tasks required for generating clinically safe and guideline-consistent antibiotic recommendations. The final convergence suggests that the KG-LLM effectively integrates knowledge-graph guidance and dental domain semantics, achieving reliable performance on this specialized medical NLP task.

*4.5 Discussion*

The experimental findings strongly demonstrate the value of integrating structured medical knowledge with large language models for pediatric dental applications. Unlike purely text-based LLMs, the KG-LLM framework benefits from explicit grounding in pharmacological guidelines and dental diagnostic ontologies, enabling the model not only to extract clinical information more accurately but also to avoid unsafe antibiotic recommendations. The significant reduction in contraindication and dosage errors shows that knowledge-guided reasoning is essential when dealing with vulnerable pediatric populations. The improvements observed in the ablation analysis further confirm that each component—knowledge graph integration, retrieval augmentation, and causal safety modeling—plays a complementary role. These results highlight that large language models, when properly constrained by domain knowledge, can move beyond pattern recognition and toward clinically trustworthy decision support. Future work may expand the model to multimodal radiographic inputs or integrate real-time clinician feedback loops to further enhance reliability and adoption in pediatric dentistry.

## 6. Conclusions

This study presents a Knowledge-Guided Large Language Model (KG-LLM) framework designed to automatically interpret pediatric dental clinical records and generate safety-aware antibiotic recommendations. Pediatric dental informatics faces persistent challenges arising from unstructured narrative notes, incomplete radiographic descriptions, and strict drug-safety constraints associated with children's rapidly changing physiology. Traditional clinical decision support systems often fail to integrate heterogeneous data sources or to ensure guideline-compliant dosing, prompting the need for a more intelligent and interpretable solution. By integrating a domain-specific knowledge graph, retrieval-augmented generation, and a dual-layer safety verification mechanism, the proposed KG-LLM advances the capabilities of dental AI systems in both diagnostic understanding and therapeutic safety.

Through comprehensive experiments on 32,000 de-identified pediatric dental visit records, the KG-LLM demonstrates strong and consistent performance gains over a domain-adapted Llama-2 clinical baseline. The model achieves notable improvements in pediatric dental record understanding (F1: 0.914 vs. 0.867), drug–dose–duration recommendation accuracy (Top-1: 0.782 vs. 0.716), and unsafe antibiotic suggestion rate (CVR: 0.042 vs. 0.084), representing a 50% reduction in potentially harmful outputs. Furthermore, the system attains superior scores in summary quality (BLEU), Top-3 accuracy, and a global clinical safety index, collectively validating its ability to generate both accurate and guideline-adherent recommendations. The ablation studies further confirm the importance of each component—knowledge graph guidance, RAG-based evidence retrieval, and causal safety checking—highlighting their synergistic contribution toward safer antibiotic decision-making in pediatric dentistry.

The findings of this research carry significant clinical and practical implications. For pediatric dentists and dental clinics, the KG-LLM provides a scalable tool capable of interpreting complex multimodal clinical records while reducing the risk of inappropriate or unsafe antimicrobial prescriptions. For healthcare institutions, it offers a transparent and evidence-grounded framework that can complement existing CDSS infrastructures and support antimicrobial stewardship initiatives. More broadly, this study demonstrates the feasibility of combining structured domain knowledge with large language models to build clinically interpretable, safety-oriented AI systems for high-risk medical scenarios.

Despite the important findings, this study has some limitations, such as [the variable reliability of external data sources and the differing levels of evidence quality across these sources. The model also has limited ability to recognize when essential patient information is missing and appropriately request clarification from the provider. In addition, it must continually adapt to the ongoing updates in medical research to ensure its knowledge remains current.] Future research could further explore [strategies to better assess and weight heterogeneous evidence sources and develop mechanisms to ensure timely, reliable knowledge updates while strengthening the model's ability to detect information gaps and prompt provider input.]

In conclusion, this study [evaluated the antibiotic recommendations produced by the proposed KG-LLM against multiple baseline models using de-identified pediatric EDRs. The results demonstrated that the KG-LLM achieved superior performance in record understanding, recommendation accuracy, and safety compliance. The ablation study confirmed the contribution of each model component, and the training loss curve showed stable learning behavior necessary for generating safe, guideline-adherent antibiotic recommendations. Overall, this proposed model offers new insights into AI-assisted record interpretation and clinical decision-making, supporting improved clinical outcomes in the management of acute pediatric dental infections.]